# Can AI Chatbots Pass the Fundamentals of Engineering (FE) and Principles and Practice of Engineering (PE) Structural Exams?

M.Z. Naser[1,2], Brandon Ross[1], Jennier Ogle[1], Venkatesh Kodur[3], Rami Hawileh[4], Jamal Abdalla[4], Huu-Tai Thai[5]

[1]School of Civil and Environmental Engineering, and Earth Sciences, Clemson University, USA
[2]Artificial Intelligence Research Institute for Science and Engineering (AIRISE), Clemson University, USA
[3]Department of Civil and Environmental Engineering, Michigan State University, USA.
[4]Department of Civil Engineering, American University of Sharjah, UAE.
[5]Department of Infrastructure Engineering, The University of Melbourne, Australia.
E-mail: mznaser@clemson.edu, Website: www.mznaser.com

The engineering community has recently witnessed the emergence of chatbot technology with the release of OpenAI *ChatGPT-4* and Google *Bard*. While these chatbots have been reported to perform well and even pass various standardized tests, including medical and law exams, this forum paper explores whether these chatbots can also pass the Fundamentals of Engineering (FE) and Principles and Practice of Engineering (PE) exams. A diverse range of civil and environmental engineering questions and scenarios are used to evaluate the chatbots' performance, as commonly present in the FE and PE exams. The chatbots' responses were analyzed based on their relevance, accuracy, and clarity and then compared against the recommendations of the National Council of Examiners for Engineering and Surveying (NCEES). Our report shows that *ChatGPT-4* and *Bard*, respectively scored 70.9% and 39.2% in the FE exam and 46.2% and 41% in the PE exam. It is evident that the current version of *ChatGPT-4* could potentially pass the FE exam. While future editions are much more likely to pass both exams, this study also highlights the potential of using chatbots as teaching assistants and guiding engineers.

**What is a chatbot?**
A Chatbot is a computer program that is designed to be capable of simulating conversation with humans (Adamopoulou and Moussiades 2020). *ChatGPT-4* and *Bard* were created and recently released by OpenAI (2023) and Google (2023). When asked to define what a chatbot is, *ChatGPT-4* and *Bard* answered:

> A chatbot is an artificial intelligence (AI) software program designed to simulate human-like conversations with users, usually through text or voice interactions. Chatbots can understand and process natural language inputs, such as text messages or spoken words, and generate appropriate responses based on their underlying algorithms and knowledge base. They are typically used to provide customer support, answer frequently asked questions, assist with various tasks, or offer personalized recommendations.
>
> Chatbots can be built using different natural language processing (NLP) techniques and machine learning algorithms, which allow them to understand and interpret the user's intent and context. Over time, chatbots can learn and improve their responses by analyzing user interactions and incorporating new knowledge into their systems. They are commonly integrated into messaging platforms, mobile applications, websites, and virtual assistants, providing users with a convenient and efficient way to access information and services.

And according to *Bard*,

> A chatbot is a computer program that simulates human conversation. It can be used to answer questions, provide information, or even complete tasks. Chatbots are often used in customer service, where they can answer questions about products or services without the need for a human representative. They can also be used in marketing, where they can generate leads or provide information about products or services.
>
> Chatbots are typically programmed using a natural language processing (NLP) system. This allows them to understand and respond to human language. NLP systems are trained on large amounts of data, such as text or transcripts of conversations. This allows them to learn how to understand and respond to human language in a natural way.

Chatbots are becoming increasingly popular as businesses look for ways to automate customer service and marketing tasks. They can be a valuable tool for businesses of all sizes, and they are likely to continue to grow in popularity in the years to come.

The same chatbots were also asked to define the Fundamentals of Engineering (FE) and Principles and Practice of Engineering (PE) exams. Below are their responses, and as one can see, these responses align well with the formal definitions of these exams, as noted by the National Council of Examiners for Engineering and Surveying (NCEES, 2023a) and others (Johnston et al. 2007).

*ChatGPT-4:*

The Fundamentals of Engineering (FE) and Principles and Practice of Engineering (PE) exams are standardized examinations administered by the National Council of Examiners for Engineering and Surveying (NCEES) in the United States. These exams are designed to assess the knowledge and skills of engineering professionals and are part of the process to obtain a Professional Engineer (PE) license.

- Fundamentals of Engineering (FE) Exam:
  The FE exam is the first step in the process of becoming a licensed professional engineer. It is designed for recent graduates and students who are close to finishing an undergraduate engineering degree from an EAC/ABET-accredited program. The FE exam is a computer-based test that covers various engineering disciplines, such as civil, mechanical, electrical, and chemical engineering. Passing the FE exam signifies that the candidate has demonstrated a fundamental understanding of engineering principles and is ready to work as an Engineer-in-Training (EIT) or Engineering Intern (EI).
- Principles and Practice of Engineering (PE) Exam:
  The PE exam is the second step in the process of becoming a licensed professional engineer. It is designed for engineers who have gained a minimum of four years of post-college work experience in their chosen engineering discipline. The PE exam tests the candidate's competency.

*Bard:*

The Fundamentals of Engineering (FE) and Principles and Practice of Engineering (PE) exams are the first and second steps, respectively, in the process of becoming a licensed professional engineer (PE). The FE exam is a computer-based exam that covers a broad range of engineering topics, while the PE exam is a more specialized exam that covers a specific area of engineering.

The FE exam is typically taken by recent graduates or students who are close to finishing an undergraduate engineering degree. The PE exam is typically taken after four years of professional engineering experience.

The FE exam is a required step for becoming a PE, but it is not a guarantee that you will pass the PE exam. The PE exam is a much more difficult exam, and it is important to be well-prepared before taking it.

There are many resources available to help you prepare for the FE and PE exams. These resources include books, online courses, and practice exams. It is important to find a resource that fits your learning style and that will help you cover the material that is on the exams.

It is also important to practice taking exams. The FE and PE exams are both multiple-choice exams, and it is important to get used to the format of the exams. There are many online practice exams available, and you can also find practice exams in books and online courses.

Taking the FE and PE exams is a challenging process, but it is a rewarding one. If you are committed to becoming a PE, then you should start preparing for the FE and PE exams as soon as possible.

**Approach and findings**
The National Council of Examiners for Engineering and Surveying (NCEES) regulates the FE and PE exams (NCEES 2023a). According to the NCEES, the FE exam is a 6-hour exam containing 110 questions, and exam takers often have 5 hours and 20 minutes to complete. The FE exam covers a variety of topics spanning 18 areas. These are listed herein, along with the expected number of questions noted by the NCEES FE civil practice exam.

On the contrary, the Principles and Practice of Engineering (PE) exam tests for a minimum level of competency in a particular engineering discipline. This exam is designed for engineers who have gained a minimum of four years of post-college work experience in a particular civil engineering discipline. The PE exam includes 80 questions and can be completed in 8 hours. This study uses the Structural Engineering discipline of the PE exam.

In our examination, we used the same questions recommended by the NCEES' practice exams available for FE and PE exam takers. Each chatbot was asked a single question at a time. Each answer is then examined against the recommended solutions provided by the NCEES FE and PE practice exam and graded separately. The grading follows a Correct/Not Correct format (i.e., without partial points), which is similar to the grading scheme used in those exams. The total number of correct and incorrect answers is compiled at the end of the process. We did not apply any curving to adjust the scores. The complete examination process took 2 hours.

Please note that the actual questions are not listed herein to maintain the NCEES' copyrights. However, a general discussion is provided to describe the outcome of this analysis. The reader is to note that both the FE and PE exams provide some visual-based questions. Some of such questions were simply transcribed to the chatbots. Overall, each chatbot was examined with 79 FE questions and 39 PE questions. While this examination does not conform to actual FE and PE exams, still the NCEES states that (NCEES 2023b),

> *NCEES practice exams help you determine what you know—and what you don't know— before taking the exam. NCEES practice exams*
> - *Contain questions from past exams*
> - *Simulate the format, style, and level of difficulty*
> - *Provide solutions or feedback on performance in each topic area*

In our analysis, we provide some samples of correct and incorrect answers as listed by the chatbots. Some of the values were slightly changed to conceal the actual questions listed by the FE and PE practice exams.

*FE exam*
The 18 areas tested at the FE exam are listed in Fig. 1. As one can see, these areas cover general engineering themes. Altogether, each area has five different questions, and a handful of areas provide more than five questions. For example, Fig. 1 shows that Geotechnical Engineering, Hydraulics and Hydraulic Systems, Mathematics, Statics, and Mechanics of Materials have the lion's share. The same figure also shows the number and distribution of questions in a typical FE exam and our analysis. It can be seen that there is good convergence among most of the areas with the exception of Surveying and Environmental Engineering – and hence aid in validating our analysis. The large variation in these two areas stems from the fact that many of the presented questions in NCEES FE practice exam can be transcribed easily (i.e., images or charts).

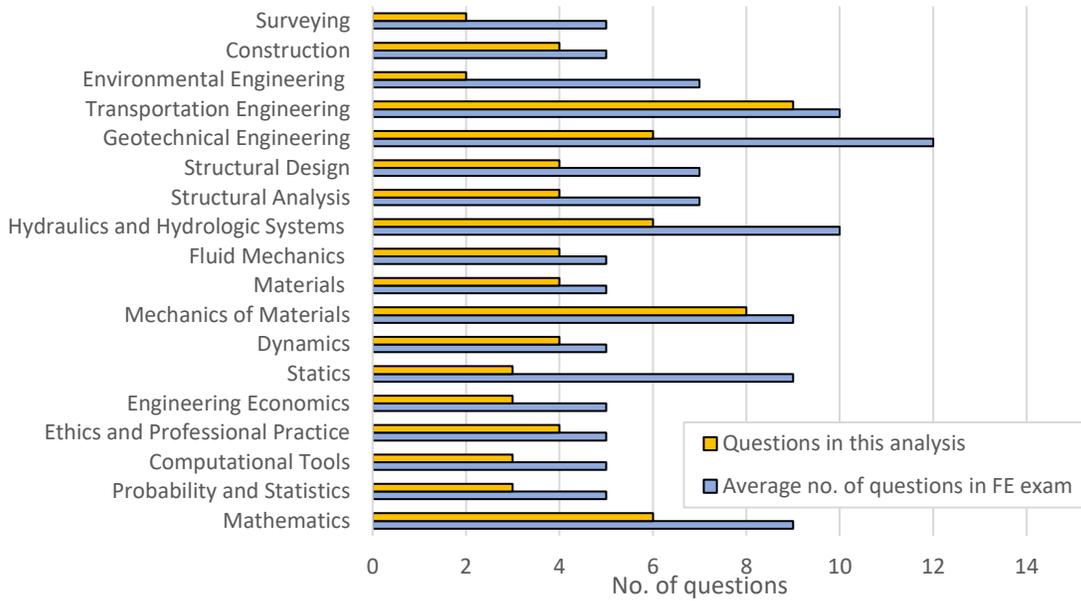

Fig. 1 Comparision between the number of questions in a typical FE exam and our analysis

Figure 2 shows the performance of *ChatGPT-4* and *Bard* across each area. It is quite clear that *ChatGPT-4* outperforms *Bard* in every single entry. In fact, *ChatGPT-4* does not perform below 50% except in Structural Design and Geotechnical Engineering areas. On the other hand, *Bard* falls short of the 50% mark consistently and does not answer any correct answers in five areas (Probability and Statistics, Computational Tools, Statics, Structural Design, and Surveying). On a different front, it is interesting to note that both chatbots do not answer any question correctly in the area of Structural Design, but they answer all questions on Ethics and Professional Practice correctly.

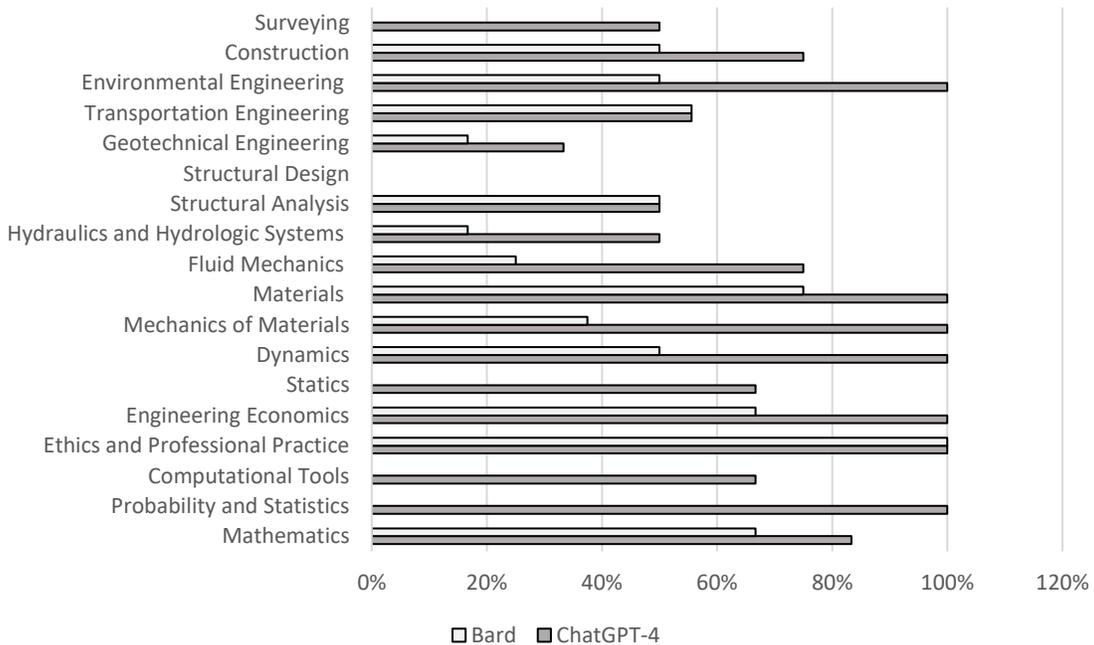

Fig. 2 Results of FE exam analysis

The findings of our analysis indicate that *ChatGPT-4* and *Bard* scored 70.9% and 39.2% in the FE exam. While the passing score for the FE exam is not commonly announced, however, such a score has been historically believed to be around 60-70%. This implies that the current version of *ChatGPT-4* could potentially pass this exam, and it is likely that future versions of *ChatGPT-4* to pass a complete and standard version of this exam as well. This rationale stems from the fact that 6 of the wrongly answered questions by *ChatGPT-4* were close to the correct answers listed by the practice exams – yet these were marked as wrong answers. For instance, in one question, *ChatGPT-4* laid out the complete process with correct formulae and logic but had a calculation mistake that skewed the final answer. In another example, *ChatGPT-4* did not convert the units properly. At the moment, it is quite interesting to note that each question addressed by *ChatGPT-4* was presented with complete thoughts and step-by-step processes and equations. Below are samples of two questions derived from those seen in the FE practice exams.

**Q.1 The area of the shaded portion of the following equation ($y = 5x^3+1$) between x = 4 and x = 9 is most nearly:**

To find the area of the shaded portion of the curve y = 5x^3 + 1 between x = 4 and x = 9, we need to integrate the function with respect to x over the given interval.

The definite integral of the function y = 5x^3 + 1 from x = 4 to x = 9 is:

∫(5x^3 + 1) dx from 4 to 9

To evaluate the integral, first find the antiderivative of the function:

Antiderivative of 5x^3 + 1 is (5/4)x^4 + x + C, where C is the constant of integration. However, since we're computing a definite integral, we don't need to worry about the constant.

Now, evaluate the antiderivative at the given limits:

[(5/4)(9^4) + 9] - [(5/4)(4^4) + 4]

Calculating the values:

(5/4)(6561) + 9 - [(5/4)(256) + 4] ≈ 8190.25 - 324 ≈ 7866.25

The area of the shaded portion between x = 4 and x = 9 is approximately 7866.25 square units.

**Q.2 The magnitude (N) of the resultant of the three coplanar forces, A, B, and C, where has an angle of 115 degrees and force of 2N, B angle of 90 degrees and force of 4.13 N, and C with angle of 5 and force of 1.15 N is most nearly:**

To find the magnitude of the resultant of the three coplanar forces A, B, and C, we can break down each force into its x and y components and then sum the components in each direction. Finally, we can calculate the magnitude of the resultant using the Pythagorean theorem.

Force A:
Magnitude: 2 N
Angle: 115 degrees

Force B:
Magnitude: 4.13 N
Angle: 90 degrees

> Force C:
> Magnitude: 1.15 N
> Angle: 5 degrees
>
> Force A components:
> Ax = 2 * cos(115°) ≈ -1.81 N
> Ay = 2 * sin(115°) ≈ 1.91 N
>
> Force B components:
> Bx = 4.13 * cos(90°) = 0 N
> By = 4.13 * sin(90°) = 4.13 N
>
> Force C components:
> Cx = 1.15 * cos(5°) ≈ 1.14 N
> Cy = 1.15 * sin(5°) ≈ 0.10 N
> Sum of x-components:
> Rx = Ax + Bx + Cx = -1.81 + 0 + 1.14 ≈ -0.67 N
>
> Sum of y-components:
> Ry = Ay + By + Cy = 1.91 + 4.13 + 0.10 ≈ 6.14 N
>
> Now, we can calculate the magnitude of the resultant force (N) using the Pythagorean theorem:
> N = √(Rx^2 + Ry^2) = √((-0.67)^2 + (6.14)^2) ≈ √(0.4489 + 37.6996) ≈ √(38.1485)
> N ≈ 6.17
> The magnitude of the resultant of the three coplanar forces A, B, and C is most nearly 6.17 N.

As one can see, the above solutions are clear and concise.

*PE exam*

Figure 3 lists the areas covered in a typical PE exam. There are 11 different areas in the Structural-based PE exam. For transparency, the examination, in the case of the PE exam, is much more complicated than that of the FE exam. This is due to the fact that the PE exam contains a larger number of questions with intricate visuals (i.e., design detailing, loading configurations, etc.) that may not be easily transcribed. As such, the number of explored questions in the presented analysis is significantly fewer than those expected in the actual PE exam (see Fig. 3). For completion, we present the findings of this analysis herein.

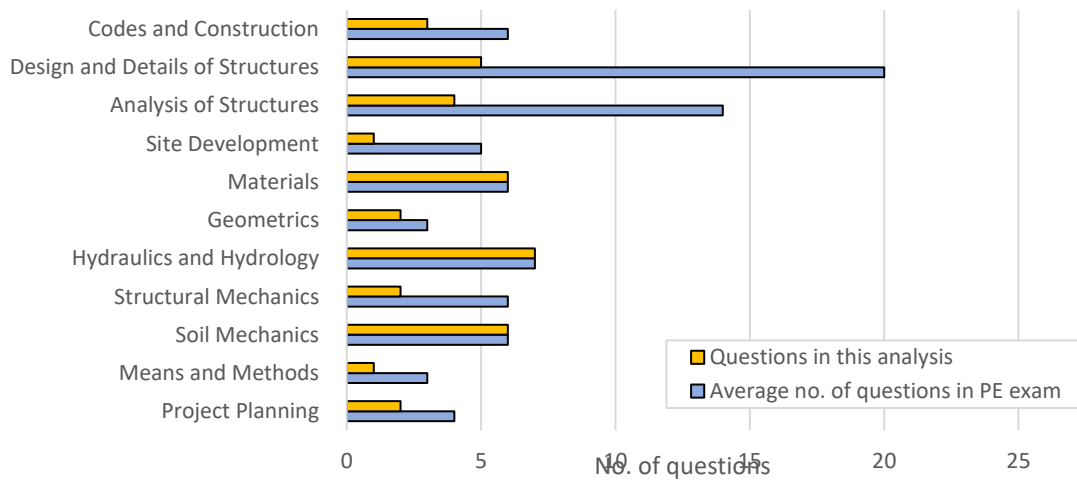

Fig. 3 Comparision between the number of questions in a typical PE exam and our analysis

Figure 4 breaks down the performance of *ChatGPT-4* and *Bard* across each area. In the case of this exam, it seems that *ChatGPT-4* and *Bard* perform equally (46.2% vs. 41%) and below the 50% mark. *ChatGPT-4* performs exceptionally well in the area of Materials and Site Development, while *Bard* performs similarly in Structural Mechanics and Site Development. In terms of Analysis of Structures, Design and Details of Structures, and Codes and Construction *ChatGPT-4* and *Bard* score within 25-67%. Unlike the case of the FE exam, the current version of *ChatGPT-4* and *Bard* are unlikely to pass this exam at the moment. We anticipate that future versions of these chatbots to be able to pass this exam.

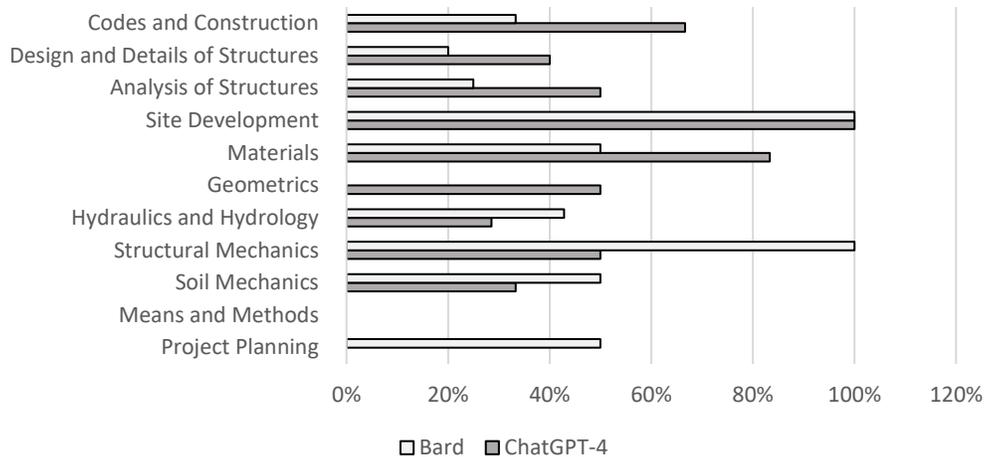

Fig. 4 Results of PE exam analysis

A few interesting observations were arrived at while examining the responses received from *ChatGPT-4* and *Bard*. For example, both chatbots seem to be able to understand the solution process behind a traditional design/analysis problem. However, the chatbots struggle to distinguish between uniform and point loads. On one occasion, *Bard* used the elastic deformation equation for a point load acting on a beam to calculate the elastic deflection of a uniform load. On the other hand, *ChatGPT-4* suffered from a simple calculation mistake (see Q. 4 below) that led to a calculation error despite having the process procedure and equations. On a more positive note, both chatbots could cite ACI and ASCE on most occasions properly. Below are samples of two questions derived from those seen in the FE practice exams.

> **Q. 3 For the following five 28-day average standard-cured concrete strength tests: A, B, C, D, and E with a required strength of G ksi, select why the concrete is unsatisfactory:**
> **(A) Because the average of three consecutive tests is less than G ksi.**
> **(B) Because Test D did not meet the required strength.**
> **(C) Because the average of the first three tests is greater than the required strength.**
> **(D) Because Test A is within 500 psi of the required strength.**
>
> According to ACI 318, the concrete is considered satisfactory if both of the following conditions are met:
> - The average of any three consecutive strength tests equals or exceeds the specified compressive strength (f'c).
> - No individual strength test result falls below the specified compressive strength by more than 500 psi.
>
> Let's evaluate the test results:

Test A psi
Test B psi
Test C psi
Test D psi
Test E psi
We check the average of three consecutive strength tests.
The average of any three consecutive strength tests equals or exceeds the specified compressive strength (f'c):
Average of Tests A, B, and C (satisfies the criterion)
Average of Tests B, C, and D (does not satisfy the criterion)
Average of Tests C, D, and e (does not satisfy the criterion)
- No individual strength test result falls below the specified compressive strength by more than 500 psi:
- Test C is 375 psi below the required strength (satisfies the criterion)

Based on these evaluations, the concrete is unsatisfactory because the average of the last two sets of three consecutive tests does not meet the required strength.
Thus, the correct statement is:
(A) Because the average of the last three consecutive tests is less than the required strength.

**Q. 4 The maximum flexural stress (psi) in a 20 ft long simply supported rectangular member with a width and height of 4 in and 12 in, loaded with a uniform load of 200 lb/ft and a point of 1000 lb at 10 ft from the support is most nearly:**

To find the maximum flexural stress, we'll first calculate the maximum moment for both the uniform load and point load and then use the flexural stress formula to find the maximum stress.

Maximum moment due to uniform load:
Uniform load, w = 200 lb/ft
Length, L = 20 ft

For a simply supported beam with a uniform load, the maximum moment (M1) occurs at the center of the span:

M1 = (w * L^2) / 8
M1 = (200 lb/ft * (20 ft)^2) / 8
M1 = 10,000 lb*ft

Maximum moment due to point load:
Point load, P = 1000 lb
Distance from support, a = 10 ft
Length, L = 20 ft

For a simply supported beam with a point load, the maximum moment (M2) is given by the formula:
M2 = (P * a * (L - a)) / L
M2 = (1000 lb * 10 ft * (20 ft - 10 ft)) / 20 ft
M2 = 5,000 lb*ft

Now, we need to find the maximum combined moment. Since both loads are on the same beam, we'll simply add the two moments to get the maximum combined moment:

M_total = M1 + M2
M_total = 10,000 lbft + 5,000 lbft
M_total = 15,000 lb*ft

> Now, we'll use the flexural stress formula to find the maximum stress:
>
> Flexural stress = (M * S) / I [Note from authors, Flexural stress = M / S]
>
> Where:
> M is the maximum moment
> S is the section modulus
> I is the moment of inertia
>
> First, we'll calculate the moment of First, we'll calculate the moment of inertia (I) for the rectangular section:
> I = (b * h^3) / 12
> Where b is the width (4 in) and h is the height (12 in).
> I = (4 in * (12 in)^3) / 12
> I = 6912 in^4 [Note from authors, the calculation mistake occurs here, I should equal 576 in^4]
>
> Now, we'll calculate the section modulus (S):
> S = I / (h / 2)
> S = 6912 in^4 / (12 in / 2)
> S = 1152 in^3
>
> Finally, we'll calculate the maximum flexural stress using the formula:
> Flexural stress = (M * S) / I [Note from authors, Flexural stress = M / S]
> Since the maximum moment is in ft, we need to convert it to inches:
> M_total = 15,000 lbft * 12 in/ft = 180,000 lbin
> Flexural stress = (180,000 lb*in) / 1152 in^3
> Flexural stress ≈ 156.25 psi
> So, the maximum flexural stress is most nearly 156.25 psi.

**Where to go from here? Chatbots as teaching assistants and engineers!**

We still would like to point out that there are critical similarities and differences between *ChatGPT-4* and a human engineer in terms of taking the exam. For example, both *ChatGPT-4* and the engineer do not have access to the internet (unlike *Bard*). The engineer, on the other hand, would have access to plenty of resources and codes. In one instance, *ChatGPT-4* stated that it does not have access to values embedded in a specific chart and hence could not solve such a question. Other items such as lucky guessing, fatigue, and stress are likely to play a toll on human behavior. Overall, the examination process for both the FE and PE took less than 2 hours. We anticipate that future *ChatGPT-4* and *Bard* generations can serve as suitable teaching assistants and guides for students and young engineers.

**Data Availability**
Some or all data, models, or code that support the findings of this study are available from the corresponding author upon reasonable request.

**Conflict of Interest**
The authors declare no conflict of interest.